\documentclass[letterpaper, 10 pt, conference]{ieeeconf}  
\usepackage{tabularx,booktabs}
\usepackage[flushleft]{threeparttable}
\usepackage{nameref}
\usepackage{amsmath}
\usepackage[utf8]{inputenc}
\usepackage{booktabs}
\usepackage{cite}
\usepackage{graphicx}
\usepackage{color}
\usepackage{url}
\usepackage{cite}
\usepackage{comment}
\usepackage{amsfonts}
\usepackage[per-mode=symbol]{siunitx}
\usepackage{mathtools}
\usepackage{setspace}
\usepackage{bm}
\usepackage{pifont}
\usepackage{algorithm,algorithmic}

\IEEEoverridecommandlockouts                              

\title{\LARGE \bf
Learning Generic and Dynamic Locomotion of Humanoids Across Discrete Terrains
}

\author{Shangqun Yu$^1$, Nisal Perera$^{1}$, Daniel Marew $^{1}$, and Donghyun Kim$^{1}$
\thanks{Authors are with the $^{1}$ Manning College of Information and Computer Sciences at University of Massachusetts Amherst, Amherst, MA, 140 Governors Dr, Amherst, MA 01002, USA. Corresponding Author: {\tt \small robot.dhkim@gmail.com}}%
}

\begin{document}

\maketitle
\thispagestyle{empty}
\pagestyle{empty}

\begin{abstract}

This paper addresses the challenge of terrain-adaptive dynamic locomotion in humanoid robots, a problem traditionally tackled by optimization-based methods or reinforcement learning (RL). Optimization-based methods, such as model-predictive control, excel in finding optimal reaction forces and achieving agile locomotion, especially in quadruped, but struggle with the nonlinear hybrid dynamics of legged systems and the real-time computation of step location, timing, and reaction forces. Conversely, RL-based methods show promise in navigating dynamic and rough terrains but are limited by their extensive data requirements. We introduce a novel locomotion architecture that integrates a neural network policy, trained through RL in simplified environments, with a state-of-the-art motion controller combining model-predictive control (MPC) and whole-body impulse control (WBIC). The policy efficiently learns high-level locomotion strategies, such as gait selection and step positioning, without the need for full dynamics simulations. This control architecture enables humanoid robots to dynamically navigate discrete terrains, making strategic locomotion decisions (e.g., walking, jumping, and leaping) based on ground height maps. Our results demonstrate that this integrated control architecture achieves dynamic locomotion with significantly fewer training samples than conventional RL-based methods and can be transferred to different humanoid platforms without additional training. The control architecture has been extensively tested in dynamic simulations, accomplishing terrain height-based dynamic locomotion for three different robots.

\end{abstract}

\section{Introduction}

\begin{figure}
\includegraphics[width=\linewidth]{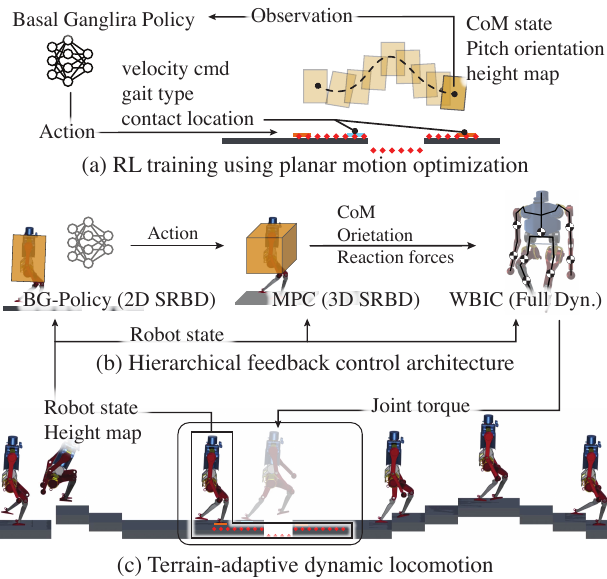}
\caption{{\bf The proposed learning framework and control architecture.} (a) We first train a policy using a single rigid body dynamics (SRBD) model in the sagittal plane and trajectory optimization. (b) The trained policy (BG-policy) is integrated into the motion controller (MPC + WBIC) to compute the final joint commands for a humanoid robot. (c) Our control architecture commands a humanoid robot to walk, leap over gaps, jump onto platforms, and navigate stairs based on vision-based data.}
    \label{fig:overview}
\end{figure}


The control of legged locomotion essentially boils down to three questions: when and where to step, and how to adjust the reaction force. A long history of significant efforts has been dedicated to finding solutions to these questions. In the early phases, researchers attempted to simplify the model by constraining certain movement dimensions, such as the linear height of the center of mass (CoM)~\cite{kajita20013d,zhao2012three}, and then tried to define a proper contact point location~\cite{vukobratovic2004zero,koolen2012capturability,englsberger2015three,goswami1999foot}. These early studies laid the foundational understanding of how legged locomotion should be analyzed and demonstrated locomotion in various robot hardware~\cite{kim2023disturbance,caron2019stair,johnson2017team}. However, the simplifications made for analysis also limited the capabilities of the controllers, typically confining the motion to conservative walking.

On the other hand, optimization-based methods, most notably model-predictive control, focus on finding optimal reaction forces. A common formulation involves solving for contact forces based on a single rigid body~\cite{cmpc, villarreal2020mpc} or centroidal momentum model~\cite{lee2007reaction}, and then computing joint commands by solving inverse kinematics or dynamics. The optimization variables can range from reaction force only~\cite{cmpc, villarreal2020mpc}, to both foot position and reaction force~\cite{bledt2017policy}, or to full joint torque based on multi-body dynamics~\cite{dantec2021whole}. These approaches have achieved highly stable and dynamic locomotion~\cite{kim2019highly,garcia2021mpc}. However, optimization-based algorithms fundamentally struggle with hybrid dynamics, making the simultaneous optimization of contact timing, location, and reaction forces (equivalently, CoM trajectory) challenging. Several contact-implicit trajectory optimization efforts have aimed to tackle this issue~\cite{werling2021fast, le2024fast, jeon2022online, kim2023contact}, but so far, no real-time optimization-based controller has succeeded in selecting both step location and timing along with reaction forces given terrain information.

Another prevailing approach is a reinforcement learning (RL)-based framework: training neural networks through interaction with the environment. Typically, the locomotion policy made with neural networks takes observations of the robot state and outputs joint commands (desired position~\cite{lee2020learning, peng2020learning,li2021reinforcement} or torque~\cite{chen2023learning, kim2023torque}). These RL-based methods bypass the three questions (i.e., step timing, location, and reaction force) by focusing on joint position commands without explicitly considering foot placement or timing. Recent strides in end-to-end RL-based methods have been impressive -- quadruped robots can now walk and run over rough terrains~\cite{lee2020learning, tan2018simtoreal, peng2020learning}, and bipeds show significant robustness and terrain adaptation~\cite{siekmann2021blind, li2021reinforcement,radosavovic2023realworld}. However, since end-to-end RL-based strategies learn from massive amounts of experience~\cite{rudin2021learning}, they require huge data sets, and incorporating increasing complexity, such as adding perception data to the locomotion, is nontrivial. Recent RL studies in perception-based locomotion required long training hours ($10\sim20~\si{\hour}$) ~\cite{cheng2023parkour,zhuang2023robot,hoeller2023anymal}. Another serious scalability issue of end-to-end style RL-based methods is that the trained policy is bound to a specific system, so small variations (e.g., longer legs, body mass changes, different joint arrangements) in a robot cannot be addressed by the policy.

In summary, both optimization-based and RL-based methods have fundamental limitations in handling terrain-adaptive dynamic locomotion. Although their challenges appear different, they stem from the fundamental limitation of gradient descent-based optimization: finding solutions becomes challenging when the search space is large and the problem is nonlinear. RL-based methods have shown better performance in handling contact decision issues due to their outstanding exploration features but require massive amounts of data. Efforts to merge these two groups have been made but the prior works often rely on heuristic algorithm to determine either contact sequence or location to simplify the problem~\cite{margolis2021learning,yu2021visual, 9779429} or opt out perception data from the observation space ~\cite{pmlr-v155-da21a, tsounis2020deepgait, xie2022glide, yang2023cajun}. To the best of our knowledge, there have been no successful demonstrations showing simultaneous selection of both contact location and sequence given terrain information. Moreover, an algorithm for perception-based dynamic locomotion involving jumping and leaping of a complete humanoid system has not been reported.

We present a new locomotion architecture consisting of a neural network policy responsible for determining high-level locomotion decisions, and a low-level optimization-based motion controller. The policy is trained through RL in a simple environment that includes only a single rigid body and contact points in planar space. In every iteration, the policy outputs the gait type (i.e., contact timing), contact location, and forward locomotion speed, then trajectory optimization (TO) with the planar model solves a convex optimization problem. In essence, the policy sees only an abstract perspective of locomotion and decides high-level decisions akin to how our basal ganglia~\cite{lanciego2012functional} operate in our brain, which is why we call our policy the Basal Ganglia-policy (BG-policy). In deployment, the trained BG-policy is combined with a low-level controller consisting of convex MPC and whole-body impulse controller (WBIC)~\cite{kim2019highly}, considered one of the state-of-the-art controllers that have demonstrated various agile behaviors in a MIT mini-cheetah~\cite{kim2019highly, jeon2022online} and humanoid robots~\cite{chignoli2021humanoid}. The proposed control architecture consisting of BG-policy, MPC, and WBIC demonstrates dynamic locomotion of humanoid robots on discrete terrains by observing the height information of upcoming terrain, selecting the proper locomotion strategies (e.g., walking, jumping, or leaping), and coordinating full-body dynamic movements.

Thanks to our efficient learning framework, we were able to train a BG-policy with only a million samples. This is significantly less than not only end-to-end based methods but also other RL-based methods utilizing optimization-based or model-based methods~\cite{Jenelten2023DTCDT,yu2021visual}. Moreover, because of the nature of the model-based motion controller, we can deploy the same BG-policy to various humanoid robots without additional training and add additional tasks other than locomotion (e.g., object carrying, head rotation). The main contributions of this paper are summarized as follows:
\begin{enumerate}
\item A novel control architecture consisting of a BG-policy and an optimal motion controller that can automatically select the proper step locations and a gait type (e.g, walking, leaping, and jumping) based on robot states and terrain height map.
\item An efficient learning framework that focuses on sagittal motion and does not require running full dynamics simulations. The trained policy is responsible for only high-level locomotion decisions, making the learning process several magnitudes more sample-efficient compared to traditional end-to-end approaches.
\item A generic locomotion controller that offers three important benefits: 1) a robot-agnostic approach allowing the policy to work on a variety of robots with zero-shot transfer, and 2) the flexibility to add extra tasks (e.g., object carrying) without additional training. 3) the capability to support omni-directional walking by seamlessly integrating an additional yaw-rate command.
\end{enumerate}

\section{Related work}
To highlight the importance of the proposed algorithm, we compiled prior works that aimed to solve the contact sequence, location, and reaction forces simultaneously. 

\subsection{Contact Implicit Trajectory Optimization (CI-TO)} 
Contact, because of its binary nature, introduces discrete jumps in the gradient when incorporated into optimization variables.  Common techniques like mixed-integer programming (MIP)~\cite{ding2018single} is suffered from exponentially growing computational time. On the other hand, contact implicit trajectory optimization (CI-TO), as initially presented in \cite{1631739} and later in \cite{posa2014direct}, use complementarity constraints to mitigate this issue. However, real-time control using CI-TO faces the risk of converging to local minima. Soft contact constraints have been suggested as a solution ~\cite{tassa2012synthesis}. More recently, advances in differential dynamic programming (DDP)-based approaches have demonstrated experimental success with quadruped robots but require starting near the solution due to DDP's local search nature ~\cite{kim2023contact,le2024fast}.
Alternatively, solving LCP with a pre-trained warm starting point generator has reduced computation time but performed poorly beyond the trained data set ~\cite{jeon2022online}. \cite{winkler2018gait} circumvented the gradient discontinuity issue by representing the contact sequence with parameterized polynomial functions. However, this method necessitates specifying the number of contact points and has not been tested in full dynamics simulations.

Currently, there is no established solution for CI-TO beyond relaxing the contact constraint. To our knowledge, CI-TO algorithms have not yet been integrated with perception-based locomotion in real-time. With this understanding, we confined the optimization problem to convex or quasi-convex forms, relegating all decision variables that introduce significant nonlinearity to the outputs of the BG-policy.


\subsection{Training efficiency of Reinforcement Learning}

Training perception-based locomotion for discrete terrains presents significant challenges, even for quadruped systems, due to the enlarged observation space and difficulties in controlling accurate step locations. For example, \cite{zhuang2023robot} required 18 hours to train depth image-based parkour locomotion on an Nvidia 3090, and \cite{miki2022learning} used a teacher-student training scheme that consumed 12 million samples for student training alone. 


The challenge intensifies when applying these end-to-end systems to humanoid robots, which likely need even more extensive training due to their larger action spaces and inherent instability. For instance, \cite{tang2023humanmimic} spent 30 hours training blind walking for humanoid robots in the Isaac Gym \cite{makoviychuk2021isaac} using an Nvidia 3090Ti. Furthermore, most studies on biped locomotion, such as \cite{pmlr-v100-xie20a, siekmann2021simtoreal, siekmann2021blind, batke2022optimizing, li2021reinforcement, radosavovic2023realworld, tang2023humanmimic}, have yet to integrate perception data. While a few studies have demonstrated the ability to learn perception-based walking on challenging terrain with biped robots like Cassie \cite{rudin2021learning, duan2023learning}, there is, to our knowledge, no existing research demonstrating similar achievements on complete humanoid systems. Additionally, several graphics community studies have shown terrain-adaptive biped locomotion \cite{2020-ALLSTEPS, singh2022learning}, but step locations are typically selected based on classical, heuristic methods rather than letting the policy choose those steps. 

In contrast, our BG-policy efficiently learns to plan dynamic locomotion for complete humanoid systems with just 1 million samples in under an hour on a laptop equipped with an 8-core AMD Ryzen 7 CPU and an RTX 3080 GPU. This efficiency is achieved despite the optimization, based on Casadi \cite{Casadi} and the Ipopt solver, running on the CPU across only 12 parallel environments. This showcases not only the potential of our approach in addressing the challenges of humanoid locomotion but also its superior efficiency compared to existing methods.

\begin{figure}
    \centering
    \includegraphics[width=\linewidth]{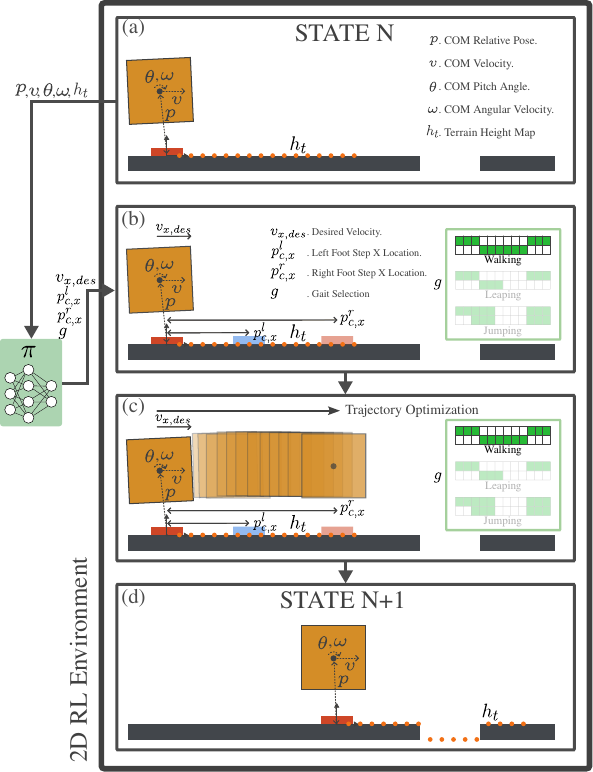}
    \caption{ {\bf State Transition in the 2D Environment.} We designed a 2D environment which makes the policy focus on only the information that matters. The simplification leads to exceptional efficient training of the policy. This environment enables the policy to be effectively trained using no more than 1 million samples, a quantity several magnitude smaller than what is typically required in end-to-end methods with vision based data.}  
    \label{fig:step}
\end{figure}

\section{Basal Ganglia Policy Training}
For efficient training, we distilled the environment down to the essential elements necessary to learn high-level decisions. Consequently, in our training environment, a humanoid robot is substituted by a planar single rigid-body and two line contacts, the left foot and right foot. Here, a line contact is formulated by paired two point contacts for implementation simplicity and the future extension to various gait types such as a heel-toe transition. The observation for the environment is
 \begin{equation}
     \mathbf{obs} = \begin{bmatrix}
         \mathbf{p}^{\top} & \theta & \mathbf{v}^{\top} & \omega & \mathbf{h_t}^{\top}
     \end{bmatrix}^{\top},
 \end{equation}
where $\mathbf{p} \in \mathbb{R}^2$ and $\mathbf{v} \in \mathbb{R}^2$ are the rigid body's center of mass position and velocity, respectively, with respect to the stance foot frame. $\theta \in \mathbb{R}$ and $\omega \in \mathbb{R}$ are the pitch angle and angular velocity, respectively, and $\mathbf{h_t} \in \mathbb{R}^{20}$ is the height map of the terrain (Fig.~\ref{fig:step}(a)). 
Our BG-policy makes an action when the right foot is under contact, where the origin of the local frame is attached to. The policy outputs the desired forward velocity, the contact locations of the next two steps (left foot and right foot) in the local frame, and the gait type (Fig.~\ref{fig:step}(b)). Subsequently, a trajectory optimisation seeks the optimal trajectory for the upcoming two steps (Fig.~\ref{fig:step}(c)) considering the contact constraints set by given action along with the current state and other constraints (e.g., reaction force and kinematics limit). The last state of the optimal trajectory, with some random Gaussian noise added, returns to the policy as the next state of the environment (Fig.~\ref{fig:step}(d)). The state of the planar single rigid body model is given by 
 \begin{equation}
     \mathbf{x} = \begin{bmatrix}
         \mathbf{p^\top} & \theta & \mathbf{v^\top} & \omega
     \end{bmatrix}^{\top},
 \end{equation}
where $\mathbf{p} \in \mathbb{R}^2$ is the position of CoM, $\theta$ is the pitch angle of the rigid body, $\mathbf{v} \in \mathbb{R}^2$ is the CoM velocity, and $\omega$ is the angular velocity. The derivative of the state $\dot{x}$ is given by
\begin{equation}
 \mathbf{\dot{x}} = \frac{d}{dt}
 \renewcommand\arraystretch{1.3}
 \begin{bmatrix}
  \mathbf{p} \\ \theta \\ \mathbf{v} \\ w
\end{bmatrix}
=
\renewcommand\arraystretch{1.3}
 \begin{bmatrix}
  \mathbf{v} \\ w \\ 
  \frac{1}{m} \sum_{i=0}^{n_c} \mathbf{f}_i - \mathbf{g} \\ 
  \frac{1}{I} \sum_{i=0}^{n_c} (\mathbf{p_c}_{i} - \mathbf{p}) \times \mathbf{f}_i
\end{bmatrix},
\end{equation}
where $\mathbf{f}_i \in \mathbb{R}^2$ is the force of the contact points. Both feet have two contact points, thus $n_c =4$ is the total number of contact points. $\mathbf{p_c}_{i} \in \mathbb{R}^2$ is the position of the $i$-th contact point. $m$ and $I \in \mathbb{R}$ are the mass and approximated inertia of a robot. To determine the value of $I$, we initially compute the inertial tensor of the robot in its nominal pose (standing position) and then project it onto the sagittal plane. The optimization is formulated by
\begin{equation*}
\min_{\mathbf{x}_k,\mathbf{f}_k} \quad  \sum_{k=1}^{N} \  (\mathbf{x}_k - \mathbf{x}^{des}_k)^\top \bm{Q_x} (\mathbf{x}_k - \mathbf{x}^{des}_k)+ \mathbf{f}_k^\top \bm{Q_f} \mathbf{f}_k
\end{equation*}
\begin{align*}
\textrm{s.t.} \quad & \mathbf{x}_{k+1} = \mathbf{x}_k + \dot{\mathbf{x}}_k \triangle t,  &\textrm{(dynamics integration)} \\[1mm]
  & -\mu f_{i,y} \leq f_{i,x} \leq \mu f_{i,y},  &\textrm{(friction)}  \\[1mm]
  & 0 \leq f_{i,y} \leq f_{max}, & \textrm{(reaction force)}\\[1mm]
  &  l_{xmin}  \leq | p_{i,x} - p_{c_i,x} | \leq l_{max, x}, & \textrm{(kinematics)} \\[1mm]
  & l_{ymin} \leq | p_{i,y} - p_{c_i,y} | \leq l_{max, x}, & \textrm{(kinematics)}\\[1mm]
  & f_{i,y} (1 - c_i) = 0, & \textrm{(contact)}
\end{align*}
where $\bm{Q_x} \in \mathbb{R}^{6\times 6 }$ and $\bm{Q_f} \in \mathbb{R}^{8\times 8}$ are weight matrices. $\mathbf{f}_k \in \mathbb{R}^{8}$ is the reaction force vector for the four contact points. $\mathbf{x}^{des}_k \in \mathbb{R}^{6}$ is the reference trajectory, which is generated by integrating the desire forward velocity output by the BG-policy. The reference trajectory's height and angle are set by nominal robot height from the ground, zero pitch angle, and zero angular velocity. $p_{i,x}$, $p_{i,y}$ represent the horizontal and vertical position of the CoM, respectively, while $p_{c_i,x}$, $p_{c_i,y}$ denotes the position of the contact point. The selection of $p_{c_i,x}$ is determined by the policy, and $p_{f_i,y}$ is determined by terrain height corresponding to $p_{c_i,x}$. 

 \begin{algorithm}[t!]
 \caption{RL Environemnt Step Function}
 \begin{algorithmic}[1]
 \renewcommand{\algorithmicrequire}{\textbf{Input:}}
 \renewcommand{\algorithmicensure}{\textbf{Output:}}\vspace{1mm}
 \REQUIRE $v_{x,des}$, $p_{c,x}^l$, $p_{c,x}^r$, $g$ 
 \ENSURE  out
  \STATE $\mathbf{p_{c}}$, $d_{violate}= $ getContactLoc($p_{c,x}^l$, $p_{c,x}^r$)\vspace{1mm}
  \STATE $\mathbf{c} =$  getContactSeqFromGait($g$)\vspace{1mm}
  \STATE $\mathbf{x}^{des}_{1..n}= $ getRef($v_{x,des}$)\vspace{1mm}
  \STATE $\mathbf{x}_{1..n}$, $l_{cost}$, $d_{terminate}=$  optimize($\mathbf{x}^{des}_{1..n}$, $\mathbf{p_c}_{,1.. n}$, $\mathbf{c}$)\vspace{1mm}
  \STATE $done = d_{violate} ||  d_{terminate}$  \vspace{1mm}
  \STATE $r=$  calcualteReward($l_{cost}$, $g$, $x_n$) \vspace{1mm}
  \STATE $\mathbf{x}_{obs}$ = $\mathbf{x}_n$ + $N(0,\sigma)$ \vspace{1mm}
  \STATE $\mathbf{h_t} = $ getHeightMap($x_{obs}$)\vspace{1mm}
 \RETURN $x_{obs}$, $\mathbf{h_t}$, $r$, $done$ 
 \end{algorithmic} 
 \end{algorithm}

The kinematics constraint are derived based on the leg length of the robot to ensure the foot step position is realizable in the actual system. $c_i \in \{0, 1\}$ indicates whether the current contact point is in contact or not, which made based on the gait selected by the policy. The reward is given by 
 \begin{equation}
     r = w_{vel} r_{vel} + w_{opt} r_{opt} + w_{gait} r_{gait},
     \label{reward}
 \end{equation}
where $w_{vel}$,  $w_{opt}$ and $w_{gait}$ are the weights for each term. The first term is a velocity reward given by 
 \begin{equation}
     r_{vel} = a  e^{- b (v_x - v_{nominal})^2},
\end{equation}
where $a>0$ and $b>0$ are hyper parameters, and $v_{nominal}$ is the nominal velocity specified by a user. The second term is the optimization reward $r_{opt} = {c}/{l}_{cost}$, where  $l_{cost}$ is the cost from the trajectory optimization, and $c >0$ is another hyper parameters. The lower the cost optimization found, the higher the $r_{opt}$ reward. The last one is the gait selection reward, to facilitate the exploration process, the reward will motivate RL to select the proper gait depending on terrain information. Building upon the described reward function, RL aims to sustain the nominal speed, minimize the resultant cost of trajectory optimization, and select an appropriate gait for diverse situations.

Algorithm 1 describe how the environment step function is designed.  The BG-policy chooses the desired forward velocity ($v_{x,des}$), horizontal step locations of the next two steps ($p_{c,x}^l$, $p_{c,x}^r$), and gait type ($g$). The environment will find the vertical step locations based on the terrain and check if the step location is feasible. For instance, if the policy decides to step in a pit then $d_{violate}$ will be set to true. The environment also gets contact sequence $\mathbf{c}$ and reference trajectory $\mathbf{x}^{des}_{1..n}$ based on $g$ and $v_{x,des}$ from the policy. Then the optimization will search for the optimal trajectory. The environment will be set to `done' if the foot step location is invalid or the optimization fail to find trajectory. The reward is then computed using Eq.~\eqref{reward}. Gaussian noise is introduced to the last state of the trajectory before it is sent as the next observation.
 


\section{Optimization-based Motion Controller}
Once the training of BG-policy is complete, we integrate the policy into the motion controller, which consists of MPC and WBIC~\cite{kim2019highly} (Fig.~\ref{fig:overview} (b)). In real-time control, we first compile the observations for the BG-policy, including the humanoid robot's CoM state, body pitch angle/velocity, and the terrain height map. Based on this information, the policy outputs the desired forward velocity, the contact locations of the left and right feet on the $x$-axis, and the gait type, and then sends them to the MPC.

\subsection{Model Predictive Control}
\begin{figure}
    \centering
    \includegraphics[width=\linewidth]{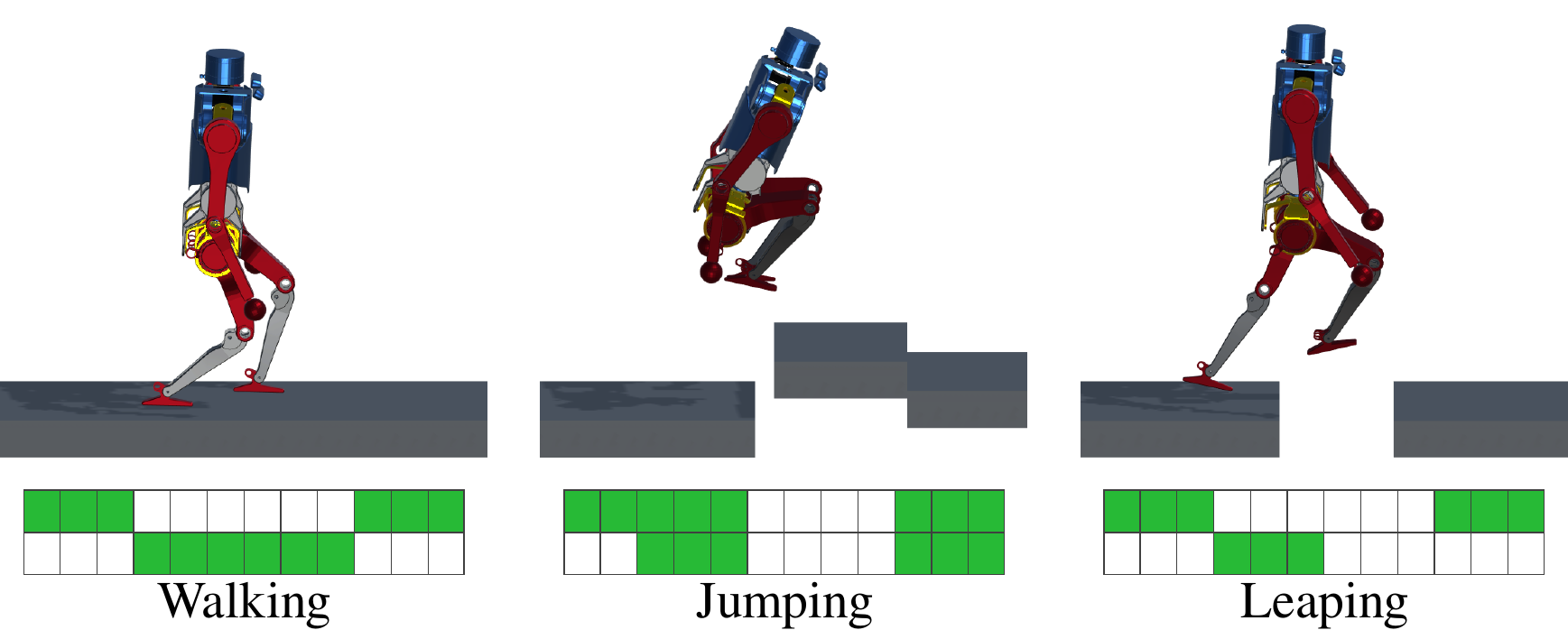}
    \caption{{\bf Three different gaits} All gaits have the same length with different contact sequence. }
    \label{fig:gait}
\end{figure}

Three gait types (i.e., walking, leaping, and jumping) are pre-specified based on appropriate swing/stance times. Each gait will lead to a different contact sequence for the next 2 steps.(Fig.~\ref{fig:gait})
Based on the high level decisions from the BG-policy, and state of a single rigid body is synchronized with the robot's CoM state and body orientation and angular velocity, then MPC calculates the reaction force, CoM/orientation trajectory, and reaction force profiles. The state of the MPC is given by 
  \begin{equation}
     \mathbf{x} = [\mathbf{p^\top} \quad \mathbf{R_{vec}} \quad \mathbf{v^\top} \quad \bm{\omega^\top}]^{\top},
 \end{equation}
where $\mathbf{p} \in \mathbb{R}^3$ and $\mathbf{v} \in \mathbb{R}^3$ are the position and the velocity of the robot's center of mass, $\mathbf{R_{vec}} \in \mathbb{R}^9$  and $\bm{\omega} \in \mathbb{R}^3$ are the vectorized orientation matrix and angular velocity of the body frame.  
The cost function and constraint are formulated in similar fashion as the 2D trajectory optimization except the orientation part. We first define $\bm{R}^{err}_k$ given by
\begin{equation}
     \bm{R}^{des}_k \bm{R}^{err}_k = \bm{R}_k, 
\end{equation} 
where $\bm{R}^{des}_k$ is the desired orientation matrix for the $k$ step and $\bm{R}_k $ is the orientation matrix at k step. Thus 
\begin{equation}
\bm{R}^{err}_k = {\bm{R}_k^{des}}^\top \bm{R}_k 
 \end{equation}
The skew-symmetric matrix form of the axis for the rotation matrix can be expressed as 
\begin{equation}
[ \hat{\bm{\omega}} ] = 
     \begin{bmatrix}
      0 & -\hat{\omega}_3 & 
 \hat{\omega}_2 \\ 
      \hat{\omega}_3 & 0 & -\hat{\omega}_1 \\
      -\hat{\omega}_2 & \hat{\omega}_1 & 0
    \end{bmatrix} 
    = \frac{1}{2 \sin{\theta}} (\bm{R} - \bm{R^\top})
 \end{equation}
 Thus 
 \begin{equation}
     \bm{\hat{\omega}}^{err}_k \theta^{err}_k = \left(\frac{\theta^{err}_k}{2 \sin{\theta^{err}_k}} (\bm{R}^{err}_k - \bm{R}^{err \top}_k)\right)^{\vee}
 \end{equation}
 $ \mathbf{\hat{\omega}_{err,k} } \theta^{err}_k \in \mathbb{R}^3$ is the orientation error, $( \cdot)^\vee : \mathfrak{so}(3) \rightarrow \mathbb{R}^3$ is the inverse of the skew function. In order to speed up the MPC, when $\theta^{err}_k$ is small, then $\theta^{err}_k \approx \sin{\theta^{err}_k}$. Our orientation error is given as 
  \begin{equation}
     \bm{\hat{\omega}}^{err}_k \theta^{err}_k = \left(\frac{1}{2} (\bm{R}^{err}_k - {\bm{R}^{err}_k}^\top)\right)^{\vee}
 \end{equation}
 The constraint for the orientation matrix is given by 
  \begin{equation}
     \bm{R}^{des}_{k+1} = \bm{R}^{des}_k ( I + [\bm{\omega}_k] \triangle t + \frac{1}{2!} [\bm{\omega}_k]^2 \triangle t^2 + \frac{1}{3!} [\bm{\omega}_k]^3 \triangle t^3 ),
 \end{equation}
 where we use the the 3rd order of the Taylor expansion to approximate the matrix exponential.  

 \begin{figure}
    \centering
    \includegraphics[width=\linewidth]{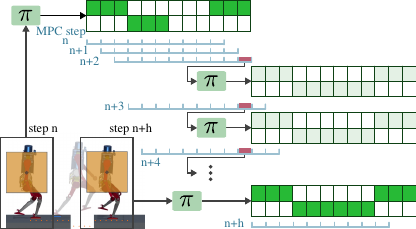}
    \caption{ {\bf Illustration of how to maintain constant prediction horizon for MPC.} Between two actions, the BG-policy  uses prediction from the MPC to compute the output, which ensure the MPC to have sufficiently long contact sequence to keep its constant prediction horizon.}
    \label{fig:rlmpc}
\end{figure}
Each action from the policy include contact sequence for 0.6 seconds, while the prediction horizon of the MPC is set at 0.5 seconds. The MPC operates in a sliding window fashion, and to maintain a constant 0.5-second window size, the policy will need to output a new sequence as soon as MPC's window hit the end of contact sequence. If the series of contact sequences from the policy do not fully cover the MPC's prediction horizon ($h$), the policy will generate a new action based on MPC's forecast (refer to Fig.~\ref{fig:rlmpc}). For instance, at step $n$, the policy has just issued an action that satisfies the MPC's prediction horizon. However, by the time it reaches step $n+3$, the policy needs to produce another sequence based on the MPC's latest state prediction, so that the MPC can maintain its horizon length. From step $n+3$ to $n+h$, the policy outputs a new action every time the MPC produces a new prediction. This approach ensures that as time progresses, the MPC's predictions become closer to the actual state, allowing the policy's actions to closely align with real-world situations. At step $n+h$, the policy can plan the contact sequence based on the actual robot state.

\subsection{Lateral Directional Step Location Selection}
While the foot step position on the sagittal plane ($x$ axis) is given by the policy, the position on the lateral direction ($y$ axis) should be calculated through the following formula: 
 \begin{equation}
     p_{f,y} = p_{y} + p_{f,y_0} + k_d v_{y},
\end{equation}
where $p_y$ is current CoM position in $y$, $p_{c,y_0}$ is the nominal relative position offset on $y$ axis. $v_y$ is the current lateral directional CoM velocity. We select $k_d$ based on velocity reversal planner from \cite{0278364920918014}, 
\begin{equation}
k_d = \sqrt{\frac{h_{\rm CoM}}{g}} \coth \left(\frac{t_{\rm swing}}{2}\sqrt{\frac{g}{h_{\rm CoM}}}\right)    ,
\end{equation}
which guarantee the asymptotic stability of linear inverted pendulum motion. The formula needs swing time ($t_{\rm swing}$) and CoM height ($h_{\rm CoM}$), and we use nominal gait parameter (swing time: $0.3~\si{\second}$, CoM height: $0.6~\si{\meter}$).
The selected number works across different gait type (walking, leaping, jumping) thanks to the planner's strong convergence. 
Once the $x$ and $y$ coordinates for the stepping position are determined, the corresponding landing height is obtained from the height map. Following this, the desired position, velocity, and acceleration of the swing foot are calculated by using a Bezier curve. 

 \subsection{Whole Body Impulse Control (WBIC)}
For the final step, WBIC~\cite{kim2019highly} is used to calculates the joint torque. WBIC utilizes a null space projection technique to build a task hierarchy which allows lower priority tasks to be executed without interfering the higher priority tasks. 
The recursion rule is given as follow.
\begin{equation}
\begin{split}
\ddot{\mathbf{q}}^{\rm cmd}_{i} &= \ddot{\mathbf{q}}^{\rm cmd}_{i-1} + \overline{\bm{J}_{i|pre}^{\rm dyn}}\left( \ddot{\mathbf{x}}_{i}^{\rm cmd} - \dot{\bm{J}}_i\dot{\mathbf{q}} - \bm{J}_{i} \ddot{\mathbf{q}}_{i-1}^{\rm cmd} \right), \\[2mm]
\label{eq:qddot_cmd}
 \end{split}
\end{equation}
\vspace{-0.5cm}
in which
\begin{align}
	&\bm{J}_{i|pre} = \bm{J}_j \bm{N}_{i-1}, \\[2mm]
\begin{split}
    &\bm{N}_{i-1} = \bm{N}_{0}\bm{N}_{1|0} \cdots \bm{N}_{i-1|i-2},\\
    &\bm{N}_{i|i-1} = \bm{I} - \overline{\bm{J}_{i|i-1}}\bm{J}_{i|i-1},\\
    &\bm{N}_{0} = \bm{I} - \overline{\bm{J}_{c}}\bm{J}_c.
    \end{split}
\end{align}
Here, $i\geq 1$, and 
\begin{equation}
    \begin{split}
        \label{eq:contact}
        \ddot{\mathbf{q}}_0^{\rm cmd} &= \overline{\bm{J}_c^{\rm dyn}}(-\bm{J}_c\dot{\mathbf{q}}).
    \end{split}
\end{equation}
$\ddot{\mathbf{x}}^{\rm cmd}_i$ is the acceleration policy of $i$-th task from the PD controller.
\begin{equation}
    \label{eq:attractor_rmp}
    \ddot{\mathbf{x}}^{\rm cmd}_i(\mathbf{x}, \dot{\mathbf{x}}) = \ddot{\mathbf{x}}^{\rm des} + \bm{K}_p \left(\mathbf{x}^{\rm des}_i - \mathbf{x}_i\right) + \bm{K}_d\left(\dot{\mathbf{x}}^{\rm des} - \dot{\mathbf{x}}\right),
\end{equation} 
where $\bm{K}_p$ and $\bm{K}_d$ are position and velocity feedback gains, respectively. $\bm{J}_{i|pre}$ is the projection of the $i$-th task Jacobian into the null space of the prior tasks. $\bm{J}_c$ is a contact Jacobian and the dynamically consistent pseudo-inverse is $\overline{\bm{J}^{\rm dyn}}$ as in~\cite{kim2019highly} and  defined as
\begin{equation}
    \overline{\bm{J}^{\rm dyn}} = \bm{A}^{-1}\bm{J}^{\top}\left( \bm{J} \bm{A}^{-1} \bm{J}^{\top} \right)^{-1}.
\end{equation}

Our task hierarchy from high to low is as follow: contact constraint $>$ body orientation task $>$ CoM position task $>$ swing foot position task $>$ joint position task. The lower priority tasks are always executed in the null-space of the higher priority tasks.
Similar to \cite{kim2019highly}, once the acceleration command $\ddot{\mathbf{q}}^{\rm cmd}_{i}$ is calculated from Eq.~\eqref{eq:qddot_cmd}, it is then tracked along side reaction force from MPC while considering the full-body dynamics of the robot by solving the following quadratic program (QP).  
After the optimal $\mathbf{f}_r$ and $\ddot{\mathbf{q}}$ are obtained by the solving the QP, inverse dynamics is used to compute the torque command. 
\begin{equation} \label{eq:qp_cost}
\min_{\bm{\delta}_{\mathbf{f}_r}, \bm{\delta}_{f}}\quad \bm{\delta}_{\mathbf{f}_r}^{\top} \bm{Q}_1 \bm{\delta}_{\mathbf{f}_r} + \bm{\delta}_{f}^{\top}\bm{Q}_2\bm{\delta}_{f}\vspace{0.7mm} \\
 \vspace{-4mm}
\end{equation}
\begin{align*}
\text{s.t.} & \\
\tag{floating base dyn.}
&\bm{S}_f 
\left(
\bm{A} \ddot{\mathbf{q}} + \mathbf{b} + \mathbf{g}
\right) 
= \bm{S}_f \bm{J}_{c}^{\top} \mathbf{f}_r \\
\tag{acceleration}
&\ddot{\mathbf{q}} = \ddot{\mathbf{q}}^{\rm cmd} +  \begin{bmatrix}
\bm{\delta}_{f} \\ \mathbf{0}_{n_j}
\end{bmatrix} \\
\tag{reaction forces}
&\mathbf{f}_r 
= \mathbf{f}_r^{\rm MPC} + \bm{\delta}_{\mathbf{f}_r}\\
\tag{contact force constraints}
 & \bm{W} \mathbf{f}_r \geq \mathbf{0},
\end{align*}
where $\bm{J}_c$ and $\bm{W}$ are the augmented contact Jacobian and contact constraint matrix respectively. $\mathbf{f}_r^{\rm MPC}$ is the reaction force command computed by the MPC, and $\bm{S}_f$ is the floating base selection matrix.  $\bm{\delta}_{\mathbf{f}_r}$  and $\bm{\delta}_{f}$ are slack variables for the reaction forces and floating base acceleration .

\section{Results}
\subsection{Experiment and Evaluation}
We use Tello robot~\cite{9813569} for the validation of our learning framework and control architecture. Firstly, we train an RL agent in the 2D environment, where the planar single rigid-body's weight, inertia and box constraint for the kinematics are extracted from the Tello robot. The algorithm we use for the BG-policy training is Soft Actor-Critic (SAC), both the actor and the critic have two hidden layers of size 256. The input layer has a size of 26, 6 for state in the sagittal plane and 20 for the height map. The resolution of the height map is $0.05~\si{\meter}$, enabling to perceive terrain features up to $1~\si{\meter}$ ahead. The output layer has 4 dimensions: desired velocity, left/right foot step, and gait selection as described in Fig.~\ref{fig:step} (b). 

In addition to the flat ground, the terrain features a variety of challenges, including gaps, high platforms, and stairs. At the start, the environment is dynamically configured by randomly selecting 10 obstacles across these categories. The distance between adjacent obstacles is also randomly determined, ranging from $1$ to $1.5~\si{\meter}$. Remarkably, the policy achieves convergence using just 1 million samples in less than an hour, running on a laptop outfitted with an 8-core AMD Ryzen 7 CPU and an RTX 3080 GPU.

\begin{figure}
    \centering
    \includegraphics[width = \linewidth]{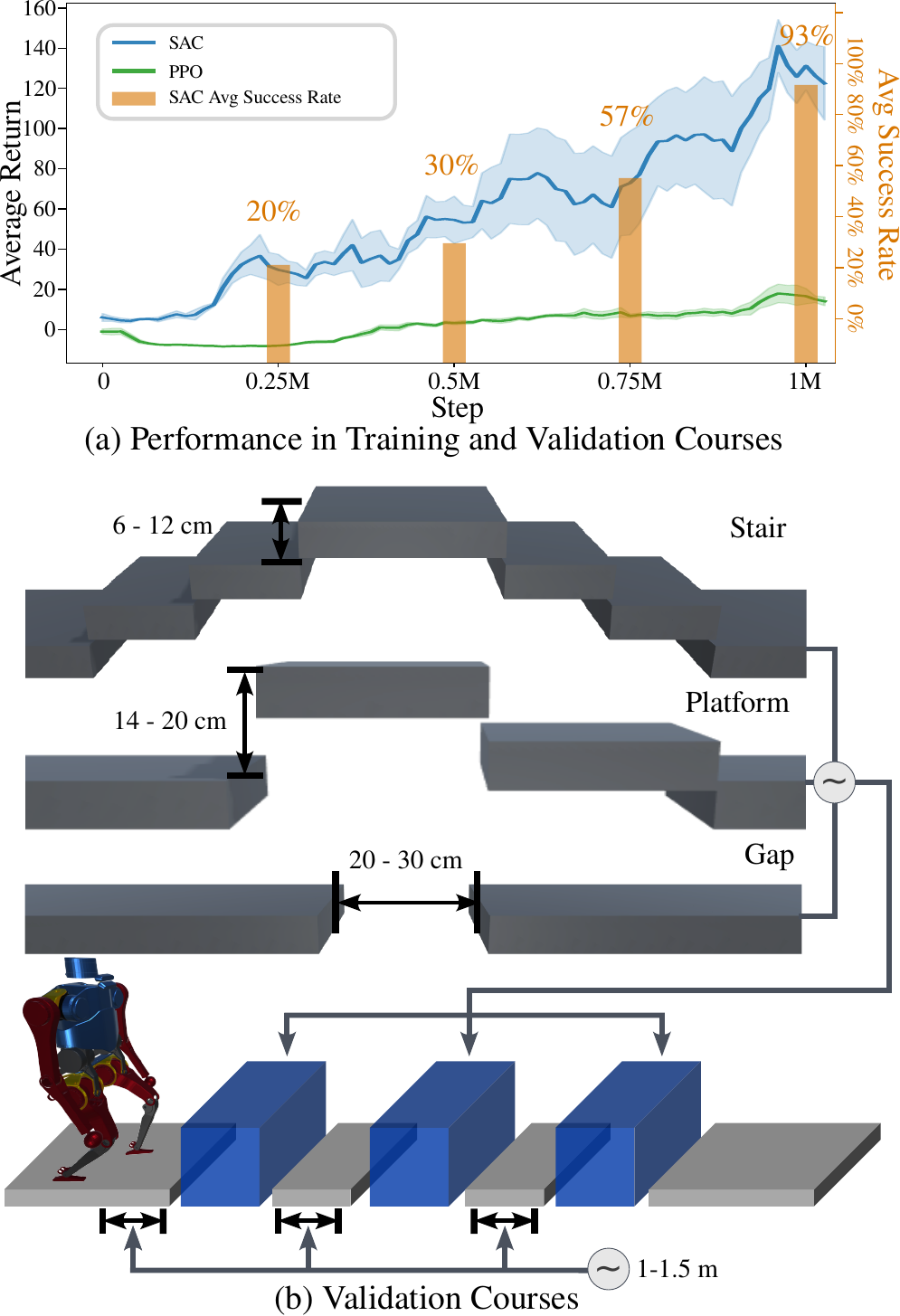}
    \caption{{\bf Training performance and Validation Courses.} (a) Both SAC and PPO are trained for 1 million steps in 5 different seeds. The average return plot shows outstanding performance of SAC algorithm. The average success rate of the policy trained by SAC is also shown as orange bar. Once the iteration reaches to 1 million steps, the policy gets converged and shows robust locomotion performance. (b) To evaluate the actual performance of the trained policy, we made 30 validation courses with randomly generated obstacles in the full dynamic simulator.}
    \label{fig:trainingProcess}
\end{figure}

The trained policy is deployed in full humanoid robot control by combining with optimization-based motion controller. In the dynamic simulation tests, we randomly generated 30 validation courses, each featuring three randomly selected obstacles with varying height/width as described in Fig.~\ref{fig:trainingProcess}(b). The 93\% success rate of SAC policy trained with 1M samples (refer Fig.\ref{fig:trainingProcess}(a)) shows that the policy can manage the dynamic navigation over discrete terrains after an hour training.

\subsection{Benchmark}
To evaluate the training efficiency depending on RL algorithms, we also tried Proximal Policy Optimization (PPO), finding that SAC significantly outperformed PPO, as shown in Fig.~\ref{fig:trainingProcess}(a). The result is also consistent with previous studies indicating that SAC, an off-policy algorithm, is more sample-efficient than on-policy algorithms like PPO, particularly in our framework that run trajectory optimization in each interaction with the environment.

Additionally, we have also compared our algorithm with the end-to-end method by recreating the same environment in the widely adopted open-source legged gym environment \cite{Nikita2021}. To verify the correctness of the implementation, the policy is firstly tested on an existing biped system (Cassie) included in \cite{Nikita2021}. Subsequently, the same end-to-end training setup is tested on the MIT Humanoid robot to evaluate the difference between a biped and a humanoid. The number of interaction steps required to achieve a 90 percent success rate in the validation courses was recorded. (Table \ref{tb:table}). Our policy proved to be 1-2 orders of magnitude more sample-efficient in terms of the number of interaction steps needed for training. While the end-to-end method with Cassie achieved a 90 percent success rate after approximately 40 million steps, the policy initially only learned to jump forward across the discrete terrain. Learning how to select the appropriate gait type for different terrains required 150 million steps. However, with the complete humanoid system, due to its high-dimensional action space, it converged to an unrealistic locomotion strategy that involved excessive torso twisting along with irregular gait sequence, and further training did not resolve this issue. One potential solution could be to employ auxiliary rewards, encouraging the humanoid to imitate an existing trajectory \cite{2018-TOG-deepMimic, 10160562, smith2023learning, batke2022optimizing, kang2023rl, tang2023humanmimic}. However, there is no existing work demonstrating vision-based locomotion on a complete humanoid using imitation learning yet, thus this approach is beyond the scope of this benchmark. 
While our algorithm demonstrates greater efficiency in terms of the number of interaction steps required with the environment, it is important to acknowledge that current GPU-based physics simulators can collect millions of samples within minutes, rendering the total time cost for the end-to-end methods comparable to ours. The primary limitation of our approach lies in the use of a CPU-based optimization solver, which restricts our ability to generate thousands of environments in parallel. However, this training time could be significantly reduced with the availability of GPU-based optimization solvers, offering a clear pathway for further improvements in efficiency.


\begin{table}
\caption{Number of step required to achieve 90\% success rate }\vspace{-0.5cm}
\begin{center}
\begin{tabular}{ |c|c|c| }
\hline
BG humanoid & E2E biped (Cassie) & E2E humanoid \\
\hline
1M & 40M & 400M\\
\hline
\end{tabular}
\label{tb:table}
\end{center}
\end{table}

\subsection{Algorithm's Robustness}
\begin{figure}
    \centering
    \includegraphics[width=\linewidth ]{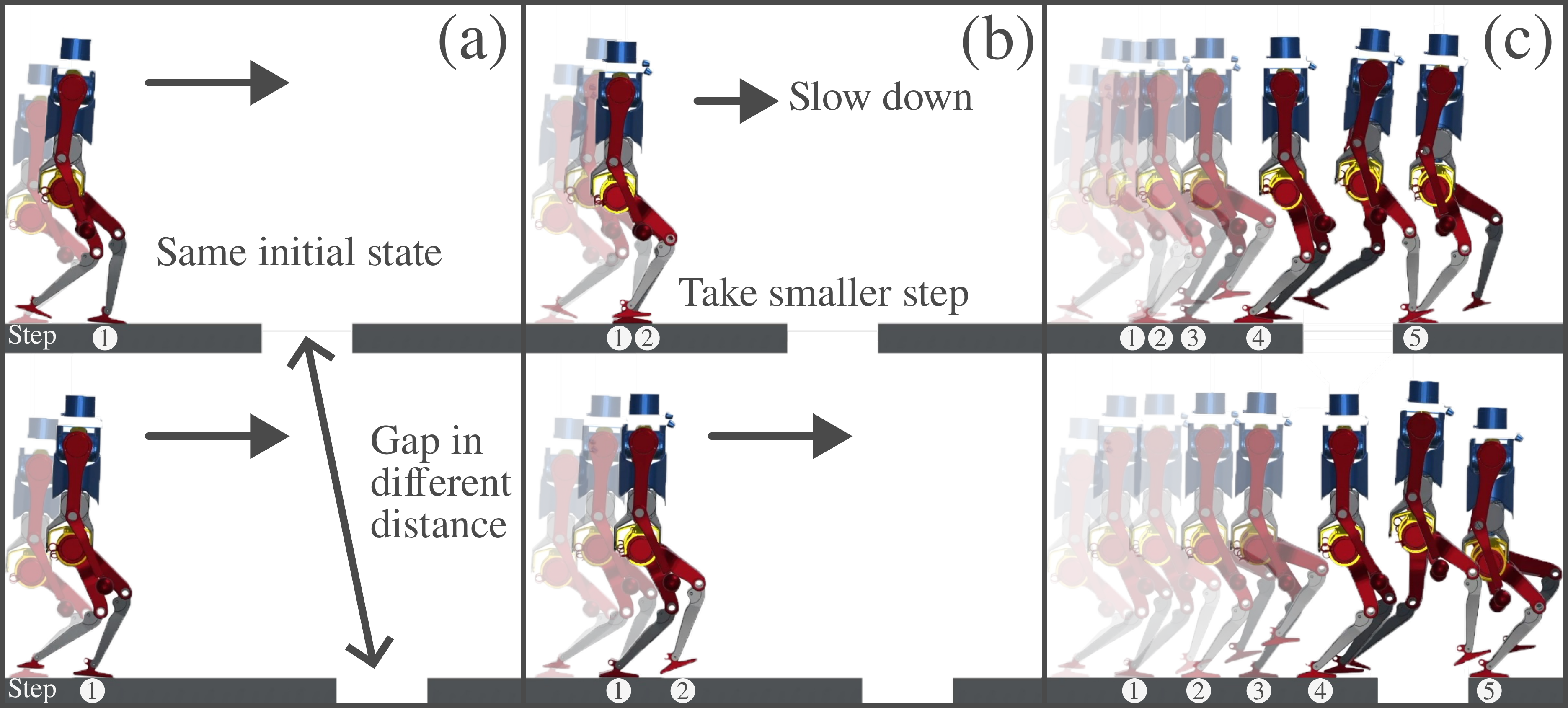}
    \caption{ {\bf Demonstrating the BG-Policy's Adaptability.} (a) The robot was initialized at the exact same state except a different distance to the gap. (b) In the top row, the policy found it need to adjust its foot step to leap over the gap, so it choose to slow down and take smaller step, while at the bottom row, the policy choose to move forward with regular step size. (c) At the end, the robot is able to leap over the gap in both situation by dynamically coordinating its forward velocity and foot step location.}
    \label{fig:stepadjust}
\end{figure}

Notably, the trained policy shows simultaneous adjustment of foot step location, gait and forward velocity to adapt to different terrain configuration. For instance, in Fig.~\ref{fig:stepadjust}, we present two distinct scenarios faced by the policy. In the scenario described in the bottom of Fig.~\ref{fig:stepadjust}, the terrain can be navigated by selecting footsteps that maintain a consistent distance from each other. However, in the scenario (shown at the top), where we deliberately reduced the gap distance from the initial position, the policy chooses to slow down, take several short steps, and then make a leap over the gap. This ability to adjust its locomotion strategy in real-time underscores the policy's effectiveness in handling diverse and challenging terrains.

\begin{figure}
    \centering
    \includegraphics[width=\linewidth]{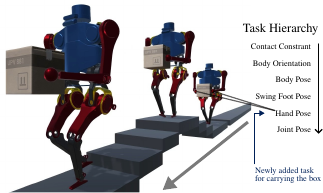}
    \caption{{\bf Tello carrying a box while traversing the irregular terrain.} By utilizing null-space projection based task prioritization, we can easily add additional task such as commanding the arm to carry a box without interfering the locomotion task, which will be very useful for object manipulation.}
    \label{fig:controlarm}
\end{figure}

Moreover, by using the WBIC, its null-space projection based task prioritization allows us to add additional task such as carrying a box (Fig.~\ref{fig:controlarm}) without interfering the existing locomotion task nor it requires additional training typically necessary in end-to-end methods.

\begin{figure}
    \centering
    \includegraphics[width=\linewidth]{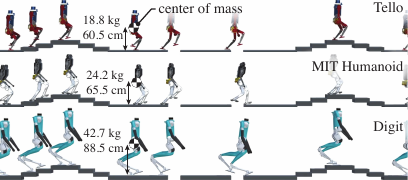}
    \caption{{\bf Locomotion of different robots using the same policy.} Since the policy is trained using an abstracted model, it can be deployed to various robots by simply switching the robot model in the motion controller. The results show that we can enable MIT Humanoid and Digit walk and jump over discrete terrain without additional training.}
    \label{fig:different_robot}
\end{figure}

Another important benefit of our framework is its flexibility, unlike the end-to-end scheme, where the trained policy is specifically tailored to the robot it was trained on. Our policy can be directly applied to other robots without requiring any additional training. We demonstrate this by applying the same policy on MIT humanoid and Digit robots, simply by switching the robot model in the low-level motion controller (Fig.~\ref{fig:different_robot}). Despite the significant difference in kinematics and dynamics between these robots and the Tello robot, for which the policy was originally trained, the same policy successfully guides different robots to navigate irregular terrain. It is noteworthy that our locomotion framework is robust to the difference between the target system being controlled and the robot on which the policy was trained. For example, although the Digit robot is significantly taller than the Tello robot, for which the policy was trained, the policy still functions effectively with a simple height adjustment of $-250~\si{\milli\meter}$ applied to the observations before they are passed into the policy.

\begin{figure}
    \centering
    \includegraphics[width=\linewidth]{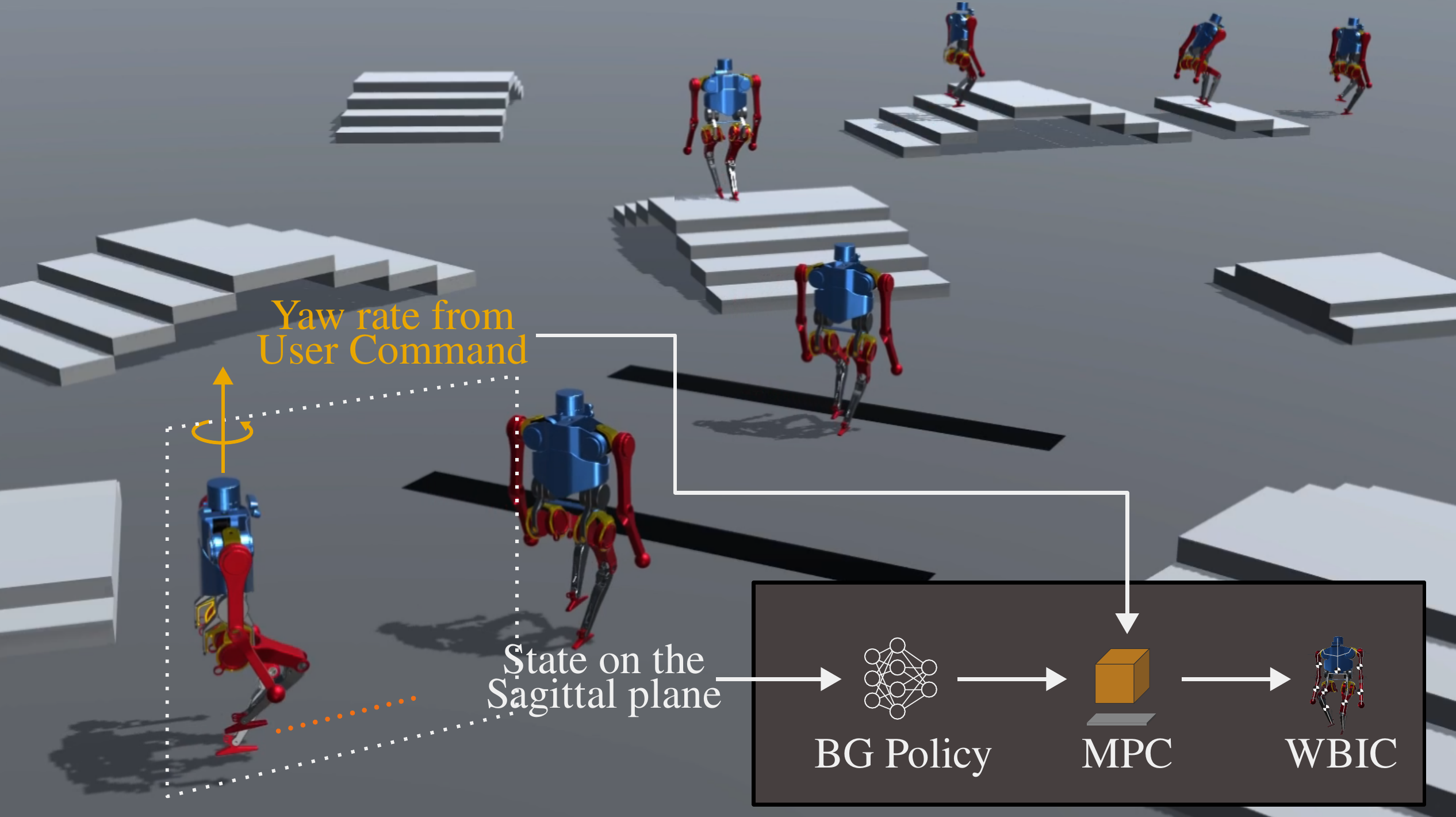}
    \caption{{\bf Demonstration of omni-directional walking.} By incorporating the user's yaw rate command, MPC+WBIC controller can manage the walking direction behavior. This rotation does not make a difference in BG-policy's perspective because it only focuses on the robot's local sagittal plane.}
    \label{fig:omni_walk}
\end{figure}

Last but not least, our algorithm also enables omni-directional walking on discrete terrain by seamlessly integrating an additional yaw rate command (Fig.~\ref{fig:omni_walk}). While taking high-level action from the BG policy, the MPC incorporates a desired yaw rate from the user command to change the walking direction. The BG policy processes state and 1D terrain information from the robot's sagittal plane -- the robot's local $xy$ plane. Then the MPC takes in the high-level actions along with the additional yaw rate command to calculate optimal trajectory and reaction force, and the WBIC calculates the joint torque commands. Because both MPC and WBIC take the robot's state in the global frame and have the complete 3D information, the algorithm does not need additional modifications to extend forward walking to omni-directional walking.

\section{Concluding Remarks}
In this work, we introduced a novel training framework and control architecture that synergizes RL-trained policy with optimal control, tailored for dynamic humanoid locomotion. Our methodology involves a streamlined 2D training environment that encapsulates critical locomotion determinants -- forward velocity, contact location, and gait selection -- leveraging terrain and state abstractions. This strategic simplification has led to a drastic reduction in the requisite training samples, surpassing traditional end-to-end approaches by orders of magnitude.
Employing this policy in tandem with low-level motion controller, our system adeptly navigates complex terrains, demonstrating a versatile locomotion repertoire that includes walking, leaping, and jumping over discrete terrains. Our control architecture facilitates the incorporation of auxiliary tasks, such as object manipulation, without compromising locomotion efficacy. Furthermore, the demonstrated zero-shot transferability underscores the policy's applicability across diverse robotic platforms. Another expected benefit is effortless sim-to-real transfer because the low-level motion controller have been successfully implemented in robot hardware.  

We also identified a limitation of the framework: when the robot's motion becomes highly dynamic, the deviation of the actual motion from the predicted trajectory becomes noticeable. Consequently, the BG-policy does not effectively accomplish what it has learned during training. Future work could explore fine-tuning the BG-policy with full dynamics simulations or integrating an additional neural network policy to bridge this gap, potentially enhancing the fidelity of motion execution
Further algorithmic enhancements could stem from leveraging GPU-accelerated optimization environments, promising substantial training speed improvements through parallelization. While our current RL setup, optimized for CPU execution, achieves significant efficiency, transitioning to GPU-based computation could further expedite the training process.

\section*{Acknowledgment}
This material is based upon work supported by the National Science Foundation under Grant No. 2220924.

\bibliographystyle{IEEEtran}
\bibliography{references}

\end{document}